\documentclass[a4paper]{llncs}
\usepackage[utf8]{inputenc}
\usepackage{cwdefs}
\usepackage{amssymb}
\usepackage{amsfonts}
\usepackage{amsmath}
\usepackage{xspace}
\usepackage{verbatim}
\usepackage{booktabs}
\usepackage[framemethod=TikZ]{mdframed}
\usepackage{tikz}
\usetikzlibrary{shapes.arrows}
\usepackage{enumitem}
\usepackage{graphicx}
\usepackage[hidelinks]{hyperref}

\newcommand{\KBSET}{\mbox{\name{KBSET}}\xspace}
\newcommand{\TEI}{\name{TEI}\xspace}
\newcommand{\LATEX}{\LaTeX\xspace}
\newcommand{\GND}{\name{GND}\xspace}
\newcommand{\SWIPL}{\name{SWI-Prolog}\xspace}

\definecolor{boxbg}{rgb}{0.92,0.92,0.92}
\definecolor{arrowbg}{rgb}{0.8,0.8,0.8}

\mdfdefinestyle{figbox}{%
     outerlinewidth=0.1pt,
    roundcorner=5pt,
    innertopmargin=7.8pt,
    innerbottommargin=7.8pt,
    innerrightmargin=7.8pt,
    innerleftmargin=7.8pt,
    backgroundcolor=boxbg}

\newenvironment{figbox}
{\begin{minipage}[t]{5.7cm}\begin{mdframed}[style=figbox]\small}
{\end{mdframed}\end{minipage}}

\newcommand{\nlfhead}{\\[3pt]}
\newcommand{\nlf}{\\[4pt]}               

\newcommand{\figarrow}{
    \begin{tikzpicture}
      \node[scale=1.0,single arrow,draw=none,fill=arrowbg,minimum height=0.7cm,shape border rotate=270] at (0,0) {};
    \end{tikzpicture}
}

\title{\KBSET\ -- Knowledge-Based Support for Scholarly Editing and Text
  Processing}

\author{Jana Kittelmann\inst{1} \and
Christoph Wernhard\inst{2}}

\institute{Martin-Luther-Universität Halle-Wittenberg, Germany\\
  \email{info@janakittelmann.de}
  \and Berlin, Germany\\\email{info@christophwernhard.com}}

\begin{document}
\maketitle

\enlargethispage{8pt}
\vspace{-12pt}
\begin{abstract}
  \KBSET supports a practical workflow for scholarly editing, based on using
  \LATEX with dedicated commands for semantics-oriented markup and a
  Prolog-implemented core system. Prolog plays there various roles: as query
  language and access mechanism for large Semantic Web fact bases, as data
  representation of structured documents and as a workflow model for advanced
  application tasks. The core system includes a \LATEX parser and a facility
  for the identification of named entities. We also sketch future perspectives
  of this approach to scholarly editing based on techniques of computational
  logic.
\end{abstract}

\section{Introduction}
\label{sec-introduction}

In the age of Digital Humanities, scholarly editing \cite{plachta} involves
the combination of natural language text with machine processable semantic
knowledge, typically expressed as markup.  The best developed machine support
for scholarly editing is the XML-based \name{TEI} format \cite{tei}, mainly
targeted at rendering for different media and extraction of metadata, achieved
through semantics-oriented or declarative markup.  Recent efforts stretch \TEI
by aspects that are orthogonal to its original \name{ordered hierarchy of
  content objects (OHCO)} text model, through support for entities like
\name{names, dates, people, and places} as well as structuring with
\name{linking, segmentation, and alignment} \cite[Chap.~13 and~16]{tei}.  Also
ways to combine \TEI with Semantic Web techniques, data modeling and
ontologies are investigated \cite{eide:2015}.  Nevertheless, there are various
demands in today's practical scholarly editing as well as with respect to
future perspectives that are not well covered by \TEI and the associated XML
processing workflow, which we will address here:

\begin{enumerate}

\item An economic workflow for scholarly editing should be supported.  Only
  very few people from the Humanities seem willing to write XML documents.
  But it should be possible for them to create, review and validate text
  annotations as well as fact bases with metadata and knowledge on entities
  such as persons and places.

\item It should be possible to generate high-quality print and hypertext
  presentations in an economic way.

\item Linking with external knowledge bases should be supported.  These
  include results of other edition projects as well as large fact bases such
  as authority files like \name{Gemeinsame Normdatei
    (GND)},\footnote{\url{http://www.dnb.de/gnd}.} metadata repositories like
  \name{Kalliope},\footnote{\url{http://kalliope-verbund.info}.}  domain
  specific bases like \name{GeoNames}, or aggregated bases like \name{YAGO}
  \cite{yago2} and \name{DBpedia} \cite{dbpedia}.

\item It should be possible to incorporate advanced semantics related
  techniques such as named entity recognition or statistics-based text
  analysis.

\item It should be possible to couple object text with associated information
  in ways that are more flexible than in-place markup: Markup can be by
  different authors or automatically generated and can be for some specific
  purpose. Queries and transformations should remain applicable also after
  changes of the markup.

\item It should be possible to associate proper logic-based semantics with
  annotations and links.  Ontology reasoning alone is not sufficient, as
  classification seems not the main operation of interest. The \GND fact base
  on persons, institutions and works, for example, gets by with a quite small
  ontology.
\end{enumerate}
Our environment \KBSET (\name{K}nowledge-\name{B}ased Support for
\name{S}cholarly \name{E}diting and \name{T}ext Processing) is on the one
  hand a practical workflow that combines different systems and is applied in
  a large project, the edition of the correspondence of philosopher and
  polymath Johann Georg Sulzer (1720--1779) with author, critic and poet
  Johann Jakob Bodmer (1698--1783).  The print version will be published as
  \cite[Vol.~10]{sulzer:gs} and, including commentaries and registers, spans
  about 2000~pages.  On the other hand, \KBSET is a prototype system that
  allows to experiment with various advanced features.

As basic format for scholarly editing \KBSET suggests to use \LATEX with a set
of newly defined custom commands that provide semantics-oriented markup
adequate for the application domain, which currently is the edition of
correspondences. This is complemented by a core system written in Prolog which
includes a \LATEX parser, an internal representation of text and annotations,
support for the representation of entities like persons, places and dates as
well as a named entity identifier based on the \GND as gazetteer.
The core version of \KBSET is available as free software from its
homepage
\begin{center}
\url{http://cs.christophwernhard.com/kbset.}
\end{center}
It comes with a demo application, the draft edition of a book from the 19th
century.  Release of the extended version of \KBSET used for the Sulzer/Bodmer
correspondence is planned together with the release of the digital edition in
the near future. Most importantly, the forthcoming version adds the
specification and support for descriptive \LATEX markup for correspondences
and supports the generation of a HTML presentation, similar to
\href{http://www.pueckler-digital.de}{\name{www.pueckler-digital.de}}
\cite{puecklerdigital}.  The 2016 version of \name{KBSET} was presented at
DHd~2016 \cite{Kittelmann:Wernhard:DHd:2016} and AITP~2016
\cite{Kittelmann:Wernhard:AITP:2016}.

The rest of this system description is structured as follows: In
Sect.~\ref{sec-workflows} we discuss the practical workflows for digital
scholarly editing supported by \KBSET.  Prolog plays various roles in the
environment, which are outlined in Sect.~\ref{sec-roles-prolog}. In
Sect.~\ref{sec-components} the Prolog-implemented core components of the
system are described.  We conclude the paper in Sect.~\ref{sec-conclusion}
with sketching future perspectives of scholarly editing and logic-based
knowledge processing.

\section{Workflows of Scholarly Editing Supported by KBSET}
\label{sec-workflows}

Three phases can be identified for machine assisted scholarly editing:
\begin{enumerate}
\item Creating the object text, enhanced by markup and other statements in
  formal languages.
  \item Generating intermediate representations for inspection by humans or
    machines, analogously to debugging.
  \item Generating consumable presentations.
\end{enumerate}
Support for all phased should be of high quality, which implies the
incorporation of existing specialized systems, in our case only free software,
in particular the GNU Emacs text editor and the \LATEX document preparation
and typesetting system along with various packages.

\begin{figure}
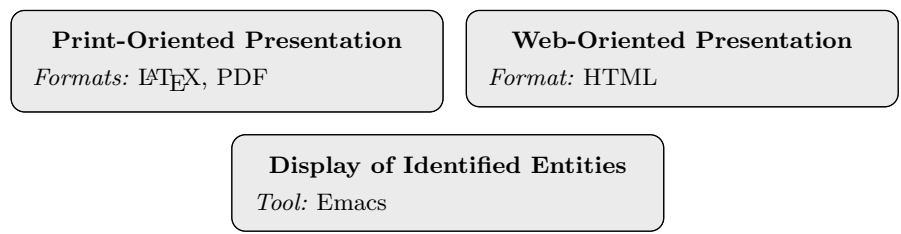

\noindent
\textbf{\large Inputs}
\vspace{-20pt}

\noindent
\hfill
\begin{figbox}
  \hfill \textbf{Object Text Documents}\hspace*{\fill}\nlfhead
  \textit{Format:} \LaTeX{} with domain specific
  semantics-based markup, e.g.\ for letter correspondences\nlf
  \textit{Tool:} Emacs
\end{figbox}
\hspace*{\fill}

\vspace{0.25cm}

\noindent
\begin{minipage}[t]{0.5\textwidth}

\hspace*{\fill}
\begin{figbox}  
  \hfill \textbf{Annotation Documents}\hspace*{\fill}\nlfhead
  \raggedright Annotations that are maintained outside of the
  object text\nlf
  \textit{Format}, \textit{Tool}: Same as for object text documents
\end{figbox}
\hspace*{0.00cm}

\vspace{0.25cm}

\hspace*{\fill}
\begin{figbox}
  \hfill \textbf{Assistance Documents}\hspace*{\fill}\nlfhead
  \raggedright
  To configure and adjust KBSET\nlf
  \textit{Format}: KBSET specific, Prolog readable\nlf
  \textit{Tool}: Emacs
\end{figbox}
\hspace*{0.00cm}
\end{minipage}
\begin{minipage}[t]{0.5\textwidth}

\hspace*{0.00cm}  
\begin{figbox}
  \hfill \textbf{Application Specific Fact Bases}\hspace*{\fill}\nlfhead
  \raggedright
  E.g. persons, works, bibliography\nlf
  \textit{Formats:} Prolog, \LaTeX{} markup, BibLaTeX\nlf
  \textit{Tools}: Emacs, JabRef
\end{figbox}

\vspace{0.25cm}

\hspace*{0.00cm}
\begin{figbox}
  \hfill \textbf{Large Imported Fact Bases}\hspace*{\fill}\nlfhead
  \raggedright
  E.g.\ GND, GeoNames, Yago, DBPedia\nlf
  \textit{Formats:} e.g. RDF/XML, CSV
\end{figbox}  
\end{minipage}

\vspace{5pt}

\hfill\figarrow\hspace*{\fill}

\vspace{-15pt}

\noindent
\textbf{\large KBSET Core System}

\noindent
\begin{minipage}[t]{0.5\textwidth}

\hspace*{\fill}
\begin{figbox}
  \hfill \textbf{Text Combination}\hspace*{\fill}
 \begin{itemize}[topsep=4pt,label=\textbullet,leftmargin=8pt]
 \item Reordering object text fragments (e.g. letters
   by different writers in chronological order)
 \item Merging with external annotations
 \item Merging with automatically generated annotations
 \end{itemize}
\end{figbox}
\hspace*{0.00pt}

\vspace{0.25cm}

\hspace*{\fill}
\begin{figbox}
  \hfill \textbf{Named Entity Identification}\hspace*{\fill}\nlfhead
  Persons, locations, dates
\end{figbox}
\hspace*{0.00pt}
\end{minipage}
\begin{minipage}[t]{0.5\textwidth}

\hspace*{0.00pt}  
\begin{figbox}
  \hfill \textbf{Consistency Checking}\hspace*{\fill}\nlfhead E.g.\ for void
  entity identifiers, insufficient or implausible date specifications, duplicate
  entries in fact bases
\end{figbox}

\vspace{0.25cm}

\hspace*{0.00pt}
\begin{figbox}
  \hfill \textbf{Register Generation}\hspace*{\fill}
  \begin{itemize}[topsep=4pt,label=\textbullet,leftmargin=10pt]
  \item Various indexes for print presentations
  \item Overview and navigation documents for Web presentation
  \end{itemize}
\end{figbox}
\end{minipage}

\vspace{5pt}

\hfill\figarrow\hspace*{\fill}

\vspace{-15pt}

\noindent
\textbf{\large Outputs}

\noindent
\begin{minipage}[t]{0.5\textwidth}

\hspace*{\fill}
\begin{figbox}
  \hfill \textbf{Print-Oriented Presentation}\hspace*{\fill}\nlfhead
  \textit{Formats:} \LaTeX, PDF
\end{figbox}
\hspace*{0.00pt}
\end{minipage}
\begin{minipage}[t]{0.5\textwidth}
\hspace*{0.00pt}  
\begin{figbox}
  \hfill \textbf{Web-Oriented Presentation}\hspace*{\fill}\nlfhead
    \textit{Format:} HTML
\end{figbox}
\end{minipage}

\vspace{0.25cm}

\noindent
\hfill
\begin{figbox}
  \hfill \textbf{Display of Identified Entities}\hspace*{\fill}\nlfhead
  \textit{Tool:} Emacs
\end{figbox}
\hspace*{\fill}

\caption{KBSET: Overview on inputs, core system
functionality and outputs}

\label{fig-overview}

\end{figure}

Figure~\ref{fig-overview} shows an overview on \KBSET.  The basic way to use
the system is the standard \LATEX workflow, however, with \LATEX commands
restricted to elements for semantics-oriented markup according to the
application domain.  For example, correspondences with letters with a sender,
recipient, date, mentioned persons, works and locations as well as scholarly
comments that are associated with specific text positions in the letters.  The
user has to manage a text editor and to know how to handle the markup
elements, which directly reflect tasks of scholarly editing.  \LATEX packages
implement these markup commands, such that the standard \LATEX workflow
immediately provides some validation and yields a formatted PDF document with
hyperlinks, realizing support for phase~(2) and also for phase~(3) with
respect to print editions.  Support to express some fact bases on entities
such as persons, works, locations and events in \LATEX syntax is supported, to
allow the user to stay in this workflow as far as possible.

Advanced functionality such as complex consistency validation, re-ordering of
document fragments such as letters and commentaries, alignment with large
external fact bases such as the \GND, automated named entity identification,
and merging with annotations that are automatically generated or maintained in
external documents, as well as conversion to other output formats like a HTML
presentation is implemented in Prolog and basically invoked through the Prolog
interpreter, although this can be hidden behind shell scripts and a GNU Emacs
interface for the named entity identification.

\begin{figure}[t]
  \centering
  \includegraphics[width=0.8\textwidth]{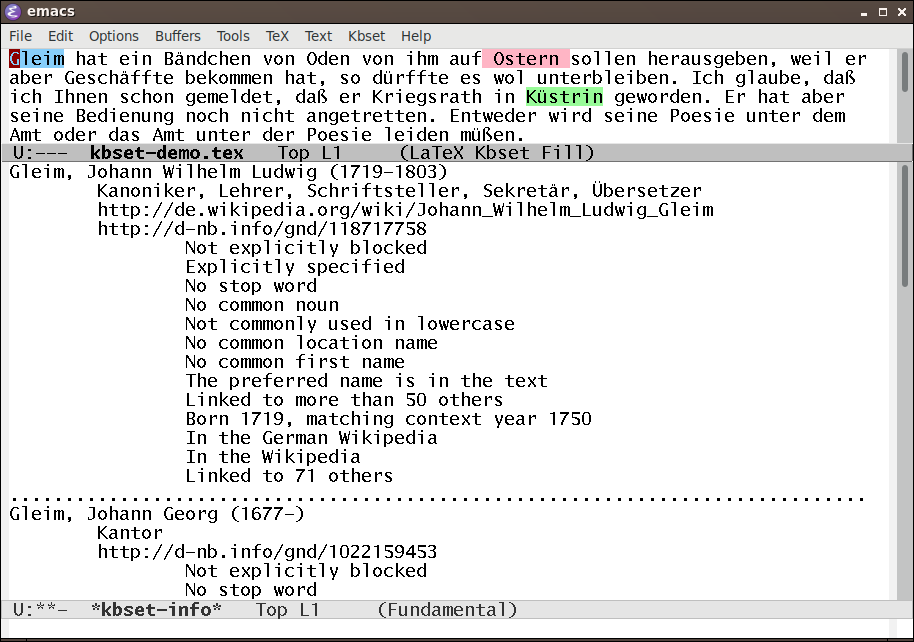}
  \caption{Screenshot: Named entity identification with \KBSET}
  \label{fig-nei}
\end{figure}

Figure~\ref{fig-nei} shows a screenshot with the presentation of named entity
identification results in Emacs.  In the object text buffer the system
highlights words or phrases about which it assumes that they denote a person,
place or date. In the lower buffer additional information on the selected
occurrence of \name{Gleim} is displayed, including a rationale for the entity
identification and a listing of lower ranked alternate candidate entities.
Further aspects of named entity identification in \KBSET will be outlined
below in Sect.~\ref{sec-nei}.

Prolog syntax is used for so-called \name{assistance documents}, that is,
configuration files where external fact bases are specified and information is
given to bias or override automated inferencing in named entity
identification. The idea is that the user, instead of annotating identified
entities manually lets the system do it automatically and mainly gives hints
in \emph{exceptional} cases, where the automatic method would otherwise not
recognize an entity correctly.  That method was used in the example document
supplied with \KBSET.  For the Sulzer/Bodmer correspondence, the primary
method was more traditional manual annotation, motivated since the mentioned
entities often need to be carefully commented anyways.  Like Prolog program
files, the assistance documents can be re-loaded, which effects updating of
the associated settings.

\section{Roles of Prolog in \KBSET}
\label{sec-roles-prolog}

The implementation language of the \KBSET core system is Prolog.  Actually,
Prolog, and in particular \SWIPL \cite{swiprolog} with its extension packages
to access modern formats like XML and RDF, is for \KBSET not just a
programming language, but covers different essential requirements within a
single system:
\begin{description}
\item[Representation Mechanism for Relational Fact Bases.] We basically
  use\linebreak \SWIPL's standard indexing facilities. Some relations are
  supplemented with semantically redundant extracts whose standard indexing
  supports specific access patterns. We call these here \name{caches}.

\item[Query Language.] The standard predicates \name{findall} and
  \name{setof} provide powerful means to specify queries in a declarative
  manner.  Complex tests and constructions can be smoothly incorporated, as
  query and programming language are identical, without much impedance
  mismatch. Problems of the interplay between different systems like difficult
  debugging and communication overhead are avoided.  Of course, queries
  written in Prolog can not rely on an optimizer, and have to be designed
  ``manually'' such that their evaluation is done efficiently.  A further
  useful feature of Prolog is sorting based on a standard order of terms. We
  used this to implement ranked answers, or top-k querying, which seems
  adequate for tasks such as searching entities that are most plausibly
  denoted by a given name.

\item[Representation Mechanism for Structured Documents.]  As in Lisp, data
  structures are in Prolog by default terms that are print- and readable, a
  feature which is supplemented to ``non-AI'' languages often as XML
  serialization.  In our application context it is particularly useful as it
  allows to represent XML and HTML documents directly as Prolog data
  structures.

\item[Parser for Semantic Web Formats.]  \SWIPL comes with powerful interfaces
  to Semantic Web formats, of which we use in particular the XML parser and
  the RDF parser, which provides a call-back interface that allows to process
  in succession the triples represented in a large RDF document (the \GND has
  about 160~million triples, the size of its RDF/XML representation is about
  2~GB).
  
\item[Workflow Model.]  Workflow aspects of experimental AI programming seem
  also useful in the Digital Humanities: loading and re-loading documents with
  formal specifications as well as invocation of functionality and running of
  experiments through an interpreter. In AI as well as in DH all of this
  should be manageable by the researcher herself instead of further parties.
\end{description}

\section{Main Components of the Prolog-Based Core System}
\label{sec-components}

Main components of the \KBSET core system are a \LATEX parser, a certain
approach to integrate large fact bases for efficient access, and a subsystem
for named entity identification that makes use of such fact bases as
gazetteers.

\subsection{\LATEX Parser}  The system includes a \LATEX parser written in
Prolog that yields a list of items, terms whose argument is a sequence of
characters represented as atom, and whose functor indicates a type such as
\name{word}, \name{punctuation}, \name{comment}, \name{command}, or
\name{begin} and \name{end of an environment}. A special type \name{opaque} is
used to represent text fragments that are not further parsed, such as \LATEX
preambles.  \LATEX commands and environments can be made known to the parser
to effect proper handling of their arguments.  The parser aims to be
practically useful, without claiming completeness for \LATEX in full.  It does
not permit, for example, a single-letter command argument without enclosing
braces. The parser is supplemented by conversions of parsing result to \LATEX
and to plain text.

\subsection{Representation of Entities from External Knowledge Bases}
\KBSET incorporates large fact bases which are typically available in Semantic
Web formats by converting them in a preprocessing phase to a set of
\name{caches}, that is, Prolog relations with extracts adapted to the
application scope (for example, retaining only data on persons born before
creation of the edited text) and access patterns required by queries (for
example, accessing a person via last name or via a \GND identifier).  These
caches can be stored in \SWIPL's \name{quick-load} format, allowing to load
them typically in a few seconds when initializing the system with application
data.  Keeping the data then in main memory does not raise problems with fact
bases such as the \GND which includes about 12 million fact triples on persons
born before 1850.  To access the relations, \KBSET supports interfaces with
predicates for entity types such as persons and locations.

\subsection{Named Entity Identifier (NEI)}
\label{sec-nei}
\KBSET includes a system for named entity identification, which detects dates
by parsing as well as persons and locations based on the \GND and
\name{GeoNames} as gazetteers, using additional knowledge from \name{YAGO} and
\name{DBpedia}.  Differently from systems like the \name{Stanford Named Entity
  Recognizer} \cite{ner:stanford}, the \KBSET NEI does not just associate
entity types such as \name{person} or \name{location} with phrases but
attempts to actually \emph{identify} the entities.  The identification is
based on single word occurrences with access to a context representation that
includes the text before and after the respective occurrence.  Hence an
association of \emph{word occurrences} to entities is computed, which is
adequate for indexes of printed documents and for hypertext presentations, but
not fully compatible with \TEI, where the idea is to enclose a \emph{phrase
  that denotes an entity} in markup.

The named entity identification is controlled by rules which can be specified
and configured and determine the evaluation of syntactic features matched
against the considered word, for example, \name{is-no-stopword} or
\name{is-no-common-sub\-stan\-tive}, and of semantic features matched against
candidate entities, for example, \name{is-in-wikipedia},
\name{is-linked-to-others-identified-in-context},
\name{has-an-occupation-mentioned-in-context}, or
\name{date-of-birth-matches-context}.  Evaluation of these features is done
with respect to the mentioned context representation, which includes general
information like the date of text creation and inferred information such as a
set of entities already identified near the evaluated text position.  Features
that are cheaply to compute and have great effect on restricting the set of
candidate entities are evaluated first.  This allows, for example, to apply
named entity identification of persons on the demo book provided with the
system, which involves several 10.000s queries against the underlying fact
bases, in about 7~seconds on a modern notebook computer.

Feature evaluation results are mapped to Prolog terms whose standard order
represent their plausibility ranking, realizing a form of top-k query
evaluation. Information about the features that contributed to selection of a
candidate entity is preserved and can be presented to the user in the form of
an explanation \emph{why} the system believes the entity to be a plausible
candidate for being referenced by a word occurrence. The Emacs interface of
\KBSET allows to browse through these candidate solutions, displaying the
explanations as well as hyperlinks to the \GND, \name{Wikipedia}, and
\name{GeoHack}, which may help to judge them (see Fig.~\ref{fig-nei} in
Sect.~\ref{sec-workflows}).  After adapting the \name{assistance document}
accordingly and re-loading it, the system will produce more accurate results
in the next run of named entity identification.

\section{Conclusion}
\label{sec-conclusion}

Digital scholarly editing involves the interplay of natural language text with
formal code and with knowledge bases in ways that suggests various interesting
possibilities related to computational logic in a long-term perspective:

There are parallels of digital scholarly editing and a classical AI scenario,
where an agent in an environment makes decisions on actions to perform, which
indicates a potential relevance of AI methods to scholarly editing: General
background knowledge in the AI scenario corresponds to knowledge bases like
\GND and \name{GeoNames}; the position of the agent in the environment
corresponds to a position in the text; temporal order of events corresponds to
the order of word occurrences; the environment which is only incompletely
sensed or understood by the agent corresponds to incompletely understood
natural language text; coming to decisions about actions to take corresponds
to decisions about denotations of text phrases and about annotations to
associate with text components.

A key requirement of a modern system to support scholarly editing is the
interplay of knowledge that is inferred by automated and statistic-based
techniques, which is inherently incomplete and not fully incorrect, with
manually supplied knowledge.  Non-monotonic reasoning should be applicable to
provide a systematic logic-based approach to mediate between the two types of
knowledge.

\KBSET already supports abstract ways to specify \name{positions} in text that
are used as target of external annotations. It seems an interesting topic of
further research to investigate this more systematically, also taking
approaches to programming into account such as the composition of information
in \name{aspect-oriented programming (AOP)} \cite{aop:97}, where items
relevant in scholarly editing roughly match concepts from AOP as follows:
Position in text -- joint point; set of positions -- pointcut; specifier of a
set of positions -- pointcut designator; action to be performed at all
positions in a set -- advice; effecting execution of ``advices'' -- weaving.

If queries are written in a suitable fragment of Prolog, they can be
automatically optimized, abstracting from caring about indexes (relation
caches), the order of subgoals and the ways in which answer components are
combined.  Recent approaches to interpolation based query reformulation might
be applicable there \cite{toman:wedell:book,benedikt:book}.  The optimized
version of a query is extracted there as a variant of a Craig interpolant from
a proof obtained from a first-order prover.  It seems also possible to apply
this approach to determine from a given set of queries the caches that need to
be constructed for efficient evaluation of the queries.

For now, we have seen with \KBSET an environment for digital scholarly editing
that has proved to be economic and practically workable in serious edition
projects. So far, the user from the Humanities applies \KBSET mainly in a
\LATEX workflow, although advanced functionality is implemented as free
software in Prolog, which is successfully and efficiently used there in a
variety of roles.

\newpage
\bibliographystyle{splncs04}
\bibliography{bib_kbset_01}

\end{document}